\documentclass{article}
\usepackage{ijcai20}

\usepackage{times}

\usepackage{soul}
\usepackage{url}
\usepackage[hidelinks]{hyperref}
\usepackage[utf8]{inputenc}
\usepackage[font=small]{caption}
\usepackage{graphicx}
\usepackage{amsmath}
\usepackage[ruled,vlined]{algorithm2e}
\usepackage{subcaption}
\usepackage{enumerate}
\usepackage{amsfonts}
\usepackage[utf8]{inputenc}
\usepackage[english]{babel}
\usepackage{amsthm}
\usepackage{balance}
\usepackage{booktabs}
\usepackage{multirow}
\newcommand{\ra}[1]{\renewcommand{\arraystretch}{#1}}
\title{Semi-Lexical Languages -- A Formal Basis for Unifying Machine Learning and Symbolic Reasoning in Computer Vision}

 \author{
 Briti Gangopadhyay$^1$\and
Somnath Hazra$^1$\and
Pallab Dasgupta$^1$\thanks{The paper is under consideration at Pattern Recognition Letters}\\
 \affiliations
 $^1$Indian Institute of Technology Kharagpur\\
 \emails
 \{briti\_gangopadhyay, pallab@cse\}@iitkgp.ac.in
 }

\begin{document}

\maketitle

\begin{abstract}
Human vision is able to compensate imperfections
in sensory inputs from the real world by reasoning
based on prior knowledge about the world. Machine
learning has had a significant impact on computer
vision due to its inherent ability in handling imprecision,
but the absence of a reasoning framework
based on domain knowledge limits its ability
to interpret complex scenarios. We propose semi-lexical
languages as a formal basis for dealing with
imperfect tokens provided by the real world. The
power of machine learning is used to map the imperfect
tokens into the alphabet of the language and
symbolic reasoning is used to determine the membership
of input in the language. Semi-lexical
languages also have bindings that prevent the variations
in which a semi-lexical token is interpreted
in different parts of the input, thereby leaning on
deduction to enhance the quality of recognition of
individual tokens. We present case studies that
demonstrate the advantage of using such a framework
over pure machine learning and pure symbolic
methods.
\end{abstract}

\section{Introduction}
Symbolic reasoning is a fundamental component of Artificial Intelligence (AI) 
which enables any rule-based system to generalize from known facts and domain 
specific rules to new facts. A necessary first step for all such systems is 
the modeling of the domain specific rules and facts in an underlying formal 
language or logic. Such systems also require the input to be encoded
in the alphabet of the language. 

One of the primary limitations of symbolic reasoning is in handling 
imperfections or noise in the system~\cite{compos}. The real world often 
presents itself imperfectly, and we require the additional ability to 
interpret the input from the real world and reduce it to the tokens in the 
alphabet. The imperfections in the input from the real world can be quite 
varied and may have individual biases, and therefore real world systems do not 
easily lend themselves succinctly to symbolic capture. Machine learning, on the other hand, is designed to handle noise in the input and thereby recognize the components of a system under various forms of imperfections. 

In this paper, we propose the notion of {\em semi-lexical languages} as the 
basis for solving several types of computer vision problems involving a 
combination of machine learning and symbolic reasoning. We accommodate
imperfections in the inputs by allowing the alphabet of the language to support
{\em semi-lexical} tokens, that is, each member of the alphabet may have many 
different variations and these variations are not defined symbolically, but 
learned from examples. For example, hand-written letters of the English 
alphabet are semi-lexical tokens. We may have many different ways in which 
people write the letter, $u$, including ways in which it may be confused with 
the letter, $v$, but we do not attempt to symbolically define all variations 
formally using more detailed features (such as the ones used by a forensic 
expert). This has the following consequences:
\begin{enumerate}
	
\item Given an input in terms of semi-lexical tokens, we need a mapping from 
    the tokens to the alphabet of the language. By the very nature of 
    semi-lexical languages, such a map is not defined symbolically, but learned
    from examples (for example, using machine learning techniques).
	
\item Depending on the level of imperfection in the semi-lexical tokens, the 
    mapping indicated above may not be unique. For example, a given 
    hand-written $u$, may be interpreted by some mapping as $u$ and by some 
    other mapping as $v$. We introduce bindings between interpretations of 
    semi-lexical tokens to ensure that the same token is not interpreted in
	two different ways if it appears multiple times in the same input. For 
	example, an individual writes the letter $u$ in a certain way, and 
	therefore, in the same sentence the hand-written letter, $u$, should not be
	interpreted in two different ways in two different portions of the text.
		
\item Since the mapping from semi-lexical tokens to the alphabet is not 
    explicit and formal, testing whether a given input is a member of the 
    language is not formally guaranteed. 
		  
\end{enumerate}
In spite of the limitation indicated in the third point above, we believe that 
semi-lexical languages are useful in representing and solving a large class of 
problems. The primary reasons are the following:
\begin{itemize}
	
\item Since the inputs from the real world often have noise and imperfections, 
    a purely symbolic form of reasoning is not possible in practice. Attempting
    to model the input variations symbolically will typically lead to 
    overfitting, and such models will not generalize to other inputs. For 
    example, different people have different ways of writing the same letters 
    and modeling the system with respect to one person's handwriting will make
    it a poor model for another person's handwriting.
	
\item Using pure machine learning is not suitable for learning complex and 
    recursively defined systems, especially when an underlying rule-based 
    structure is known and can be reduced to practice.
		  
\end{itemize}
As an example, consider the problem of training a neural network to learn the 
{\em less than} relation among digits by training it with hand-written digits. 
Machine learning is good at learning to recognize the hand-written digits~\cite{mnistsurvey}, but 
in the absence of the knowledge of the number system, the neural network will 
have to be explicitly trained for each pair of digits. It will not be able to 
generalize, for example, to deduce $3 < 7$ even when it has been trained with 
$3 < 5$ and $5 < 7$ ~\cite{evans:2018}. A semi-lexical approach, as proposed in
this paper, will use machine learning to learn the hand-written digits and use 
a back-end algebraic rule-based system to decide whether a given
input, such as $9 < 3$, is correct.

In this paper we consider two interesting case studies combining computer 
vision and symbolic reasoning to demonstrate the use of semi-lexical languages.
\begin{itemize}
	
\item The first case study examines a hand-written solution of a Sudoku puzzle 
    where some of the digits are ambiguous. The task is to decide whether the 
    solution is valid. We use this case study as a running example.
		
\item The second case study develops a framework for recognizing bicycles in 
    images. Machine learning is used to learn the components and symbolic 
    spatial constraints are used to decide whether the components add up to a 
    bicycle. We demonstrate the advantage of this approach over methods which
	train a neural network to recognize bicycles as a whole.

\end{itemize}
It is important to separate our work from previous structured component-based 
approaches such as stochastic AND/OR graphs, and from the proponents of using 
machine learning as a front-end of GOFAI\footnote{GOFAI stands for \em Good Old
Fashioned AI}, though the notion of semi-lexical languages subsumes such 
approaches. This paper includes a section on related work for this purpose.

The paper is organized as follows. Section ~\ref{sec2} formalizes the notion of semi-lexical languages, sections ~\ref{sec3} and ~\ref{sec4} elaborate the case studies. Section ~\ref{sec5} presents an overview of the related work. Section ~\ref{sec6} provides
concluding remarks.

\section{Semi-Lexical Languages}
\label{sec2}
Formally, a semi-lexical language, ${\cal L} \subseteq {\Sigma}^*$, is defined using the following:
\begin{itemize}
	
	\item The alphabet, $\Sigma$, of the language
	
	\item A set of rules (or constraints), ${\cal R}$, which defines the membership of a word
		$\omega \in {\Sigma}^*$ in the language, ${\cal L}$.
	
	\item Semi-lexical domain knowledge in the form of a set ${\cal T}$ of tagged semi-lexical
		tokens. Each semi-lexical token, $t$, is tagged with a single tag, {\em Tag(t)}, where
		{\em Tag(t)} $\in \Sigma$. We refer to ${\cal T}$ as the {\em training set}.
	
	\item A set, ${\cal C}$, of semi-lexical integrity constraints.		

\end{itemize}
In order to elucidate our proposal of semi-lexical languages, we shall use a running case study 
for the game of Sudoku, a Japanese combinatorial number-placement puzzle. The objective of the 
game is to fill a $9 \times 9$ grid with digits so that each column, each row, and each of the
nine $3 \times 3$ subgrids that compose the grid contain all of the digits from 1 to 9.

\begin{figure*}[!t]
\centering
  \begin{tabular}[b]{cc}
    \begin{tabular}[b]{l}
      \begin{subfigure}[b]{0.60\linewidth}
        \includegraphics[scale=0.128]{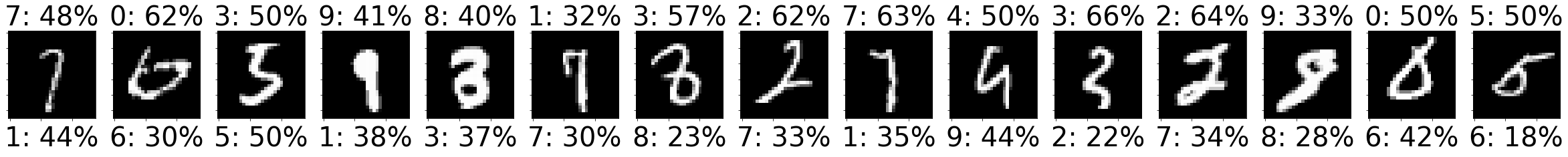}
        \caption{}
        \label{fig1:a}
      \end{subfigure}\\
      \begin{subfigure}[b]{0.38\linewidth}
        \includegraphics[scale=0.18]{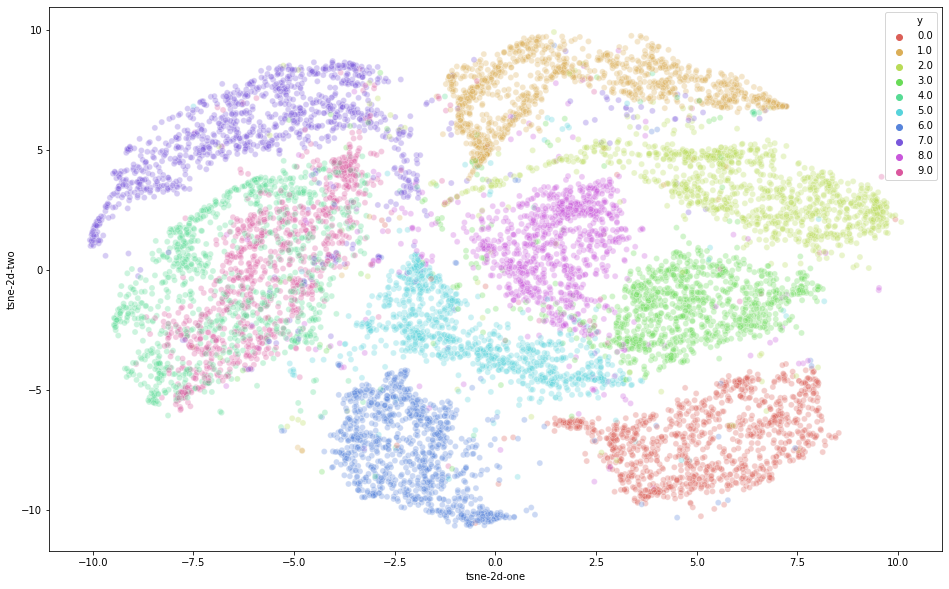}
        \caption{}
        \label{fig1:b}
      \end{subfigure}
      \begin{subfigure}[b]{0.22\linewidth}
        \includegraphics[width=1.45in,height=1.45in]{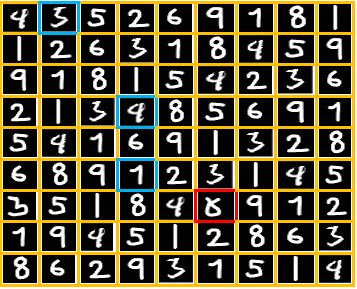}
        \caption{}
        \label{fig1:c}
      \end{subfigure}
    \end{tabular}
    &
    \begin{subfigure}[b]{0.34\linewidth}
      \includegraphics[scale=0.37]{Mapping2.png}
      \caption{}
      \label{fig1:d}
    \end{subfigure}
  \end{tabular}
  \label{fig:CycleDetection}
  \caption{a) Some semi-lexical tokens from MNIST dataset along with global support for top 2 classes b) t-SNE plot of 1000 random MNIST digits shows some digits are extremely close to others in their 2D latent representation c) Sudoku board with highlighted semi lexical tokens. d) Mapping of selected tokens from the board in Figure \ref{fig1:c} to $\Sigma$, the edges are marked with global support, $C_{2,3}$ is both globally and locally consistent where as $C_{7,6}$ is locally inconsistent.}
\end{figure*}

Let $C_{ij}$ denote the entry in the $i^{th}$ row and $j^{th}$ column of the Sudoku table.
Formally, the language, ${\cal L}$, defining the valid solutions of Sudoku is as follows:
\begin{itemize}
	
	\item The alphabet, $\Sigma = \{ 1, \ldots, 9 \}$
	\item  \label{const:2} We consider words of the form: $\omega = R_1 \|\| \ldots \|\| R_9$,
		where $R_i$ represents a row of the Sudoku, that is: $R_i = C_{i,1} \ldots C_{i,9}$. 
		A given word $\omega$ belongs to ${\cal L}$ only if it satisfies the following set
		${\cal R}$ of constraints for all $i, j$:
		\begin{enumerate}
			\item $C_{i,j} \in \{1,...,9\}$
			\item $C_{i,j} \neq C_{i',j'}$ if $i'=i$ or $j'=j$, but not both
			\item $C_{i,j} \neq C_{i',j'}$ if $\lfloor i/3 \rfloor = \lfloor i'/3 \rfloor$
				and $\lfloor j/3 \rfloor = \lfloor j'/3 \rfloor$, but not $(i=i')\land(j=j')$
		\end{enumerate}
		The second constraint enforces that no two elements in a row or column are equal, and
		the third constraint enforces that no two elements in each of the $3 \times 3$ subgrids
		are equal.
	\item The set ${\cal T}$ of semi-lexical tokens consists of various handwritten images of the
		digits. The t-SNE plot in Figure \ref{fig1:b} of 1000 random handwritten digits from MNIST dataset \cite{mnist} show that some digits like 9 and 4 are extremely close to each other in their latent representation exhibiting semi lexical behaviour. Each image is tagged with a member of $\Sigma$, that is, a digit 
		from $1, \ldots, 9$.
	\item A set, ${\cal C}$, of semi-lexical integrity constraints, which is elaborated later .		
		
\end{itemize}
Let us now consider the problem of determining whether a string of semi-lexical tokens is 
recognized as a word of the language. In the Sodoku example, our input is a $9 \times 9$ table
containing handwritten digits. The inherent connotation of semi-lexical languages allows the
tokens present in the input to be outside the training set ${\cal T}$ as well. As opposed to
formal languages, the set of semi-lexical tokens is potentially infinite. For example, there
may be infinite variations in the way people write a given letter.

%  \begin{figure*}
%   \subcaptionbox{\label{fig1:a}}{\includegraphics[scale=0.18]{tsne.png}}\hspace{\fill}
%   \subcaptionbox{\label{fig1:b}}{\includegraphics[width=1.45in,height=1.4in]{BoardSemi.png}}
%   \hspace{\fill}%
%   \subcaptionbox{\label{fig1:c}}{\includegraphics[scale=0.38]{Mapping1.png}}
%          \caption{a) t-SNE plot of 1000 random MNIST digits shows some digits are extremely close to others in their 2D latent representation exhibiting handwritten digits are semi lexical b) Sudoku board with highlighted semi lexical tokens. c) Mapping of selected tokens from board in Figure b to $\Sigma$, the edges are marked with global confidence and for $C_{76}$ the local support is mentioned.}
%          \label{fig:three graphs}
% \end{figure*}

Let ${\cal SLT}$ denote the (potentially infinite) set of semi-lexical tokens from
the real world. Obviously ${\cal T} \subseteq {\cal SLT}$. To determine whether a word 
$\omega \in {\cal SLT^*}$ belongs to ${\cal L}$, we require a mapping:
\[ {\cal F}: {\cal SLT} \rightarrow \Sigma \]
A naive way to look at semi-lexical languages would be to use machine learning (such as a 
convolutional neural network) to learn the mapping ${\cal F}$ from the tagged training set,
${\cal T}$ and then use that mapping on ${\cal SLT}$. Such an approach would have
the following pitfalls, specifically when deciding tokens which are ambiguous (similar to
more than one member of $\Sigma$). We use the Sudoku example to explain.
\begin{enumerate}

	\item {\em Inconsistent Penalties}. In Figure~\ref{fig1:c}, $C_{1,2}$ is interpreted by 
			${\cal F}$ as the digit $5$, whereas interpreting it as the digit $3$ would have
			yielded a valid solution. 

	\item {\em Inconsistent Rewards}. In Figure~\ref{fig1:c}, $C_{7,6}$ is interpreted by
			${\cal F}$ as the digit $6$ and the solution is found to be valid. However in
			$C_{2,3}$ the digit $6$ is written in a completely different way, and the same
			person is unlikely to write the digit $6$ in these two different ways.

\end{enumerate}
In human cognition, the systems of vision and reasoning support each other. We see some
parts of an object, deduce other parts of it from domain knowledge, and this deduction
is used as additional evidence in recognizing the other parts of the object which may not 
be visible with the same clarity. Our aim is to develop such methods with semi-lexical
languages as the basis.

The pitfalls indicated above can be addressed by adding {\em integrity constraints} on
the mapping ${\cal F}$ from semi-lexical tokens to the alphabet $\Sigma$, and making the 
mapping a part of the underlying reasoning system. In other words, the support for mapping
a semi-lexical token to a member of the alphabet comes from two sources, namely support
from the learning based on the training set ${\cal T}$, and support from the evidence
provided by the reasoning system which tests membership of the entire word in the 
language. Broadly we categorize the integrity constraints, ${\cal C}$, into two types:
\begin{enumerate}
	\item {\em Reasoning Assisted Similarity Constraints}. The main idea here is that 
		the rules in ${\cal R}$ can be used in conjunction with semi-lexical tokens of
		low ambiguity to hypothesize the interpretation of the ambiguous tokens. The
		hypothesis acts as increased support for interpreting the ambiguous tokens in 
		a certain way.

	\item {\em Reasoning Assisted Dissimilarity Constraints}. The main idea here is that
		two semi-lexical tokens which are very different should not be allowed to be
		mapped to the same member of the alphabet if they appear in the same word.  

\end{enumerate}
As of now, we refrain from formalizing the definition of an integrity constraint any further,
because we realize that the nature of such constraints will be very domain-specific and
susceptible to the level of noise in the training data and input. We shall demonstrate the
use of such types of constraints through our case studies.

\section{Handwritten Sudoku}
\label{sec3}
The broad steps of our semi-lexical approach towards validating a
handwritten Sudoku board is outlined in Algorithm~\ref{algo:1}. The
given image is segmented to extract the images of the digits in each
position of the board. These are then mapped to the digits $1$ to $9$
using the CNN and the board is validated using the rules of Sudoku. The 
semi-lexical analysis becomes apparent when some of the images are
ambiguous, which is reflected by low support from the CNN, and 
justifies the need for our semi-lexical approach for reasoning about such
images. We elaborate on this aspect in the following text. 
\begin{enumerate}

\item We use a CNN with only two convolution layers followed by 
    max-pooling, fully connected and softmax activation layers to 
    learn handwritten digits using the MNIST dataset. $Tag(C_{ij})$ 
    denotes the digit recognized by the CNN at position $C_{ij}$.
   
\item In order to formalise integrity constraints for handwritten digits 
    we use two distance based metrics with respect to the training data 
    $\mathcal{T}$ and local handwritten digits present on the board.
\begin{eqnarray}
f_{gs}(\mathcal{T}_t) & = &  top_k(\|(g^{l-1}(\mathcal{T}_t),g^{l-1}(\mathcal{T})\|) \label{eq1:a}
\\
f_{ls}(\mathcal{T}_t) & = & 
\frac{1}{n}\sum _{i=1}^n{\frac{1}{m}\sum_{j=1}^m{fdist_j(\mathcal{T}_t,S_{i})}}
\label{eq1:b}
\end{eqnarray}

    The function $g^{l-1}$ computes representations from the penultimate layer of 
    the neural network in order to capture translation invariance provided by the 
    maxpool layer. For each instance $\mathcal{T}_t \in $ $\mathbb{R}^{784}$, 
    $g^{l-1}$ gives a representation $\mathcal{T}'_t \in $ $\mathbb{R}^{128}$ (our architecture has 128 neurons).
    Token $\mathcal{T}_t$ is {\em globally consistent} if the confidence for the 
    correct class in the top $k$ neighbours calculated using L2 norm from 
    $g^{l-1}(\mathcal{T})$ (using Equation~\ref{eq1:a}) is greater than a lower 
    confidence bound $c_l$, that is, $f_{gs}(\mathcal{T}_t)$ $\geq c_l$.

    To calculate {\em local consistency} we check the average feature 
    distance ($fdist$) over $m$ common features calculated via {\em scale-invariant
    feature transform} (SIFT)~\cite{sift} over all $n$ similar tokens $S$ on the board (using 
    Equation~\ref{eq1:b}). Token $\mathcal{T}_t$ is {\em locally consistent} if 
    $f_{ls}(\mathcal{T}_t)$ $\leq \epsilon$.

\item The cells $C_{ij}$ with $f_{gs}(Tag(C_{ij})) \geq c_h$ are assigned the predicted 
    $Tag(C_{ij})$, otherwise the location is treated as blank. In our experiments we
    used a $c_h$ of 80\%. The $valid(Board)$ function checks whether the board satisfies
    the Sudoku constraints, $\cal R$. If not then the blank positions are solved using
    backtracking using $\cal R$ and the reasoning assisted constraints as outlined
    below.

\item The function, $GlobalSupport()$, in Algorithm~\ref{algo:1} uses function $f_{gs}$
    to compute the $k$ nearest member neighbours of the token in $C_{ij}$. It then 
    generates a $support\_map$ defining confidence for each alphabet that the token 
    image shows membership in. For example the image in $C_{4,4}$ in 
    Figure~\ref{fig1:c} has $430$ members of class $4$ and $460$ members of class $9$ 
    having similar last layer activations. Therefore 
    $support\_map(C_{4,4}) = \{ 9:46\%, 4:43\% \}$. In our experiments we used $k=1000$.

\item The blank positions representing the ambiguous digits in the board may be 
    completed using reasoning, but only without violating the reasoning assisted 
    similarity / dissimilarity constraints. The constraints are represented as a 
    bipartite graph $G = \langle V, E \rangle$ where 
    $V = V_X \cup V_Y$, $V_X = \{C_{i,j} \}$ and $V_Y = \{1,\ldots,9\}$.  
    The edges $E \subseteq V_X \times V_Y$ are determined using $f_{gs}$. An edge 
    $\langle C_{i,j}, m \rangle$ exists in $G$ iff $support\_map(C_{i,j})$ for digit $m$
    is more than the lower confidence bound $c_l$. In our experiments, we used
    a $c_l$ of 10\%. Figure~\ref{fig1:d} shows the graph $G$ for the board of 
    Figure~\ref{fig1:c}. The edges in the graph enable {\em reasoning assisted
    similarity} by virtue of multiple edges incident on a vertex of $V_X$.
    The objective is to choose $C_{ij} \rightarrow \Sigma_i$ $\mid$  $f_{gs}(C_{ij}) \geq c_l$ \& $f_{ls}(C_{ij})$ $\leq \epsilon$.This is achieved by the bipartite graph.

\item The function $Solve(board, support\_map)$ is used to choose an edge incident on each $C_{ij}$ of the bipartite graph $G$. {\em Reasoning assisted dissimilarity
    constraints} are used while making this choice. For example, $C_{7,6}$ has 
    membership in both $6$ and $8$ (that is, $\langle C_{7,6}, 6 \rangle \in E$ and
    $\langle C_{7,6}, 8 \rangle \in E$). In the absence of reasoning assisted
    dissimilarity, $\langle C_{7,6}, 6 \rangle$ may be chosen. However, the average 
    SIFT feature distance over all $8$ cells containing $6$ in the board is 
    $LocalSupport(C_{7,6}) = 11.59$, whereas $LocalSupport(C_{2,3}) = 5.49$, $\epsilon = 10$ in our experiment. This implies that the cell $C_{7,6}$ does not match with other tokens on the board having similar tag and it should not be allowed to map to the same vertex of $V_Y$ as $C_{2,3}$. The
    function Solve returns a valid board iff it is able to map each vertex of
    $V_X$ without violating any of the reasoning assisted dissimilarity constraints. 

\end{enumerate}
We highlight the fact that $7$ written in cell $C_{6,6}$ has membership in both $7$ and 
$1$, and can therefore be interpreted as $1$. Training the learning system to fit these
variations would lead to overfitting. Reasoning assisted correction overcomes this 
shortcoming of assuming pure learning-based predictions to be correct.

\begin{algorithm}[t]
\SetAlgoLined
\DontPrintSemicolon
\caption{ \small Semi-Lexical Validation of Handwritten Sudoku}
\label{algo:1}
\KwIn{$BoardImage$, $N$}
\SetKwFunction{FMain}{Main}
\SetKwProg{Fn}{Function}{:}{}
\Fn{\FMain{$BoardImage$, $N$}}{
    \For{each $image$ in $BoardImage$}{
        $Tag(C_{i,j}) \gets CNN(image)$\;
        \eIf{$f_{gs}(Tag(C_{i,j}))$ $\geq 80\%$}{
            $board[i][j] \gets Tag(C_{i,j})$\;
        }
        {
            $board[i][j] \gets blank$\;
        }
    }
    \eIf{$valid(board)$}{
        \KwRet{No Ambiguities}\;
    }
    {
        \For{each $blank$ $C_{i,j}$ in $Board$}{
            $suppport\_map \gets GlobalSupport(C_{i,j})$\;
        }
        $board \gets Solve(board, suppport\_map)$\;
        \eIf{$valid(board)$}{
            \KwRet{Corrected Board}\;
        }
        {
            \KwRet{Not Solvable}\;
        }
    }
 }
%vspace{-5cm}
\end{algorithm}

\section{Uni/Bi/Tri-Cycle Identification Problem}
\label{sec4}
Many real world vision problems have more abstract constraints than the Sudoku example.
In this section we consider one of the more popular problems, namely that of 
identifying different types of cycles. We define the alphabet as 
$\Sigma = \{wheel, seat, frame, handlebar\}$. The following rule $\mathcal{R}$ defines a bi-cycle. 
\begin{tabbing}
aaaa \= aaa \= aa \= aa \= \kill
$\exists w_1, \exists w_2, \exists f,\ C1 \land C2$, where:\\
C1: \> $wheel(w_1) \land wheel(w_2) \land w_1 \neq w_2 \land$ \\
\>\>  $\forall w_3,\ wheel(w_3) \Rightarrow (w_1 = w_3) \lor (w_2 = w_3)$ \\
C2: \> $\exists f,\ frame(f) \land inrange(f, w_1, w_2) \land$\\
\>\> $\forall f',\ frame(f') \Rightarrow (f' = f)$
\end{tabbing}
These constraints express that a bi-cycle must have two distinct wheels $w_1$ 
and $w_2$ (constraint C1), and a single frame, $f$, which is spatially within the 
range of both the wheels (constraint C2). The rules for defining uni-cycles and 
tri-cycles are similarly encoded. 
\begin{figure*}[!t]
\centering
  \begin{tabular}[b]{ccc}
      \begin{subfigure}[b]{0.15\linewidth}
        \includegraphics[scale=0.08]{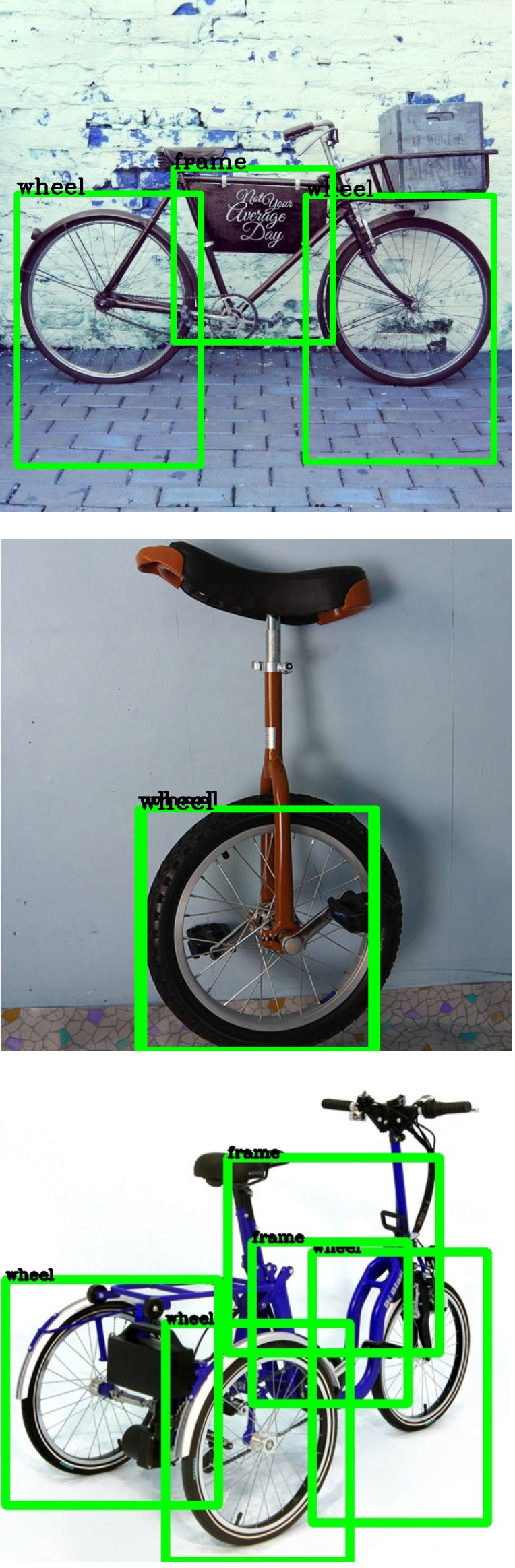}
        \caption{}
        \label{fig2:a}
      \end{subfigure}
      &
      \begin{tabular}[b]{c}
          \begin{subfigure}[b]{0.29\linewidth}
            \includegraphics[scale=0.3]{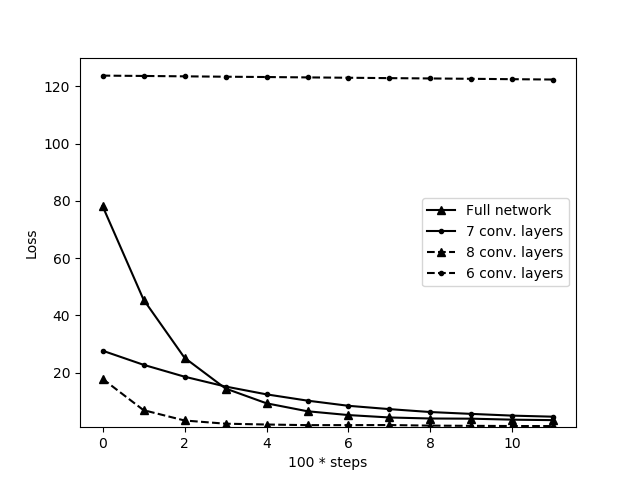}
            \caption{}
            \label{fig2:b}
          \end{subfigure}\\
          \begin{subfigure}[b]{0.24\linewidth}
            \includegraphics[scale=0.22]{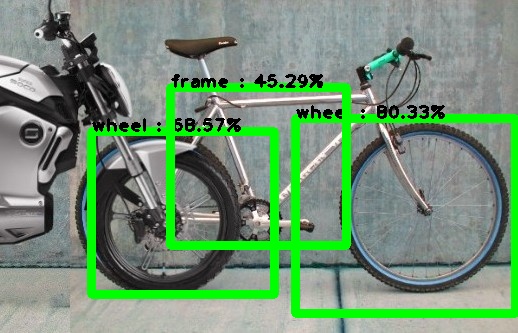}
            \caption{}
            \label{fig2:c}
          \end{subfigure}
    \end{tabular}
    \begin{subfigure}[b]{0.4\linewidth}
      \includegraphics[scale=0.1]{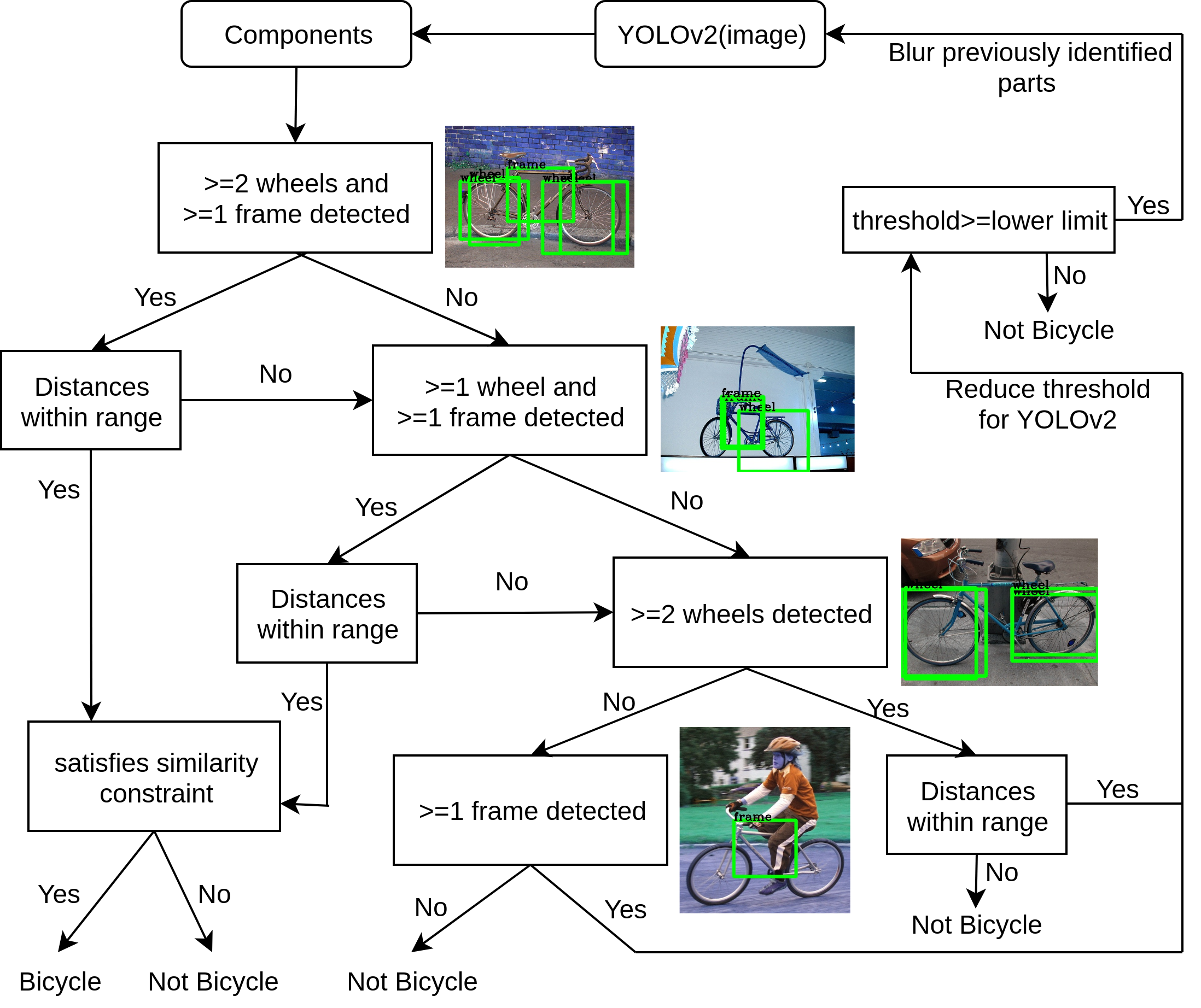}
      \caption{}
      \label{fig2:d}
    \end{subfigure}
  \end{tabular}
  \label{fig:CycleDetection}
  \caption{a) Detected components wheels and frame for different object classes b) Loss curves comparing networks with 9, 8, 7 and 6 convolutional layers, the network with 7 layers is chosen. c) Wheel of motorcycle detected to be that of bicycle however, it is flagged to be inconsistent following integrity constraints d) A decision diagram illustrating rules for identifying an object as bicycle}
\end{figure*}

The predicates, $wheel()$, $frame()$, and
$inrange()$ will have semi-lexical connotations, For example, the association of a wheel to uni/bi/tri-cycle can be ambiguous if the prediction is made only in terms of features. In the proposed semi-lexical framework the membership will therefore be resolved based on rules. As opposed to studies on stochastic AND-OR graphs,
and other shape grammars, the rules will be used to enhance the interpretation of
the vision by using the reasoning assisted knowledge to resolve ambiguities.

The symbolic rules can be used to enforce a decision chain, as shown in 
Figure~\ref{fig2:d}. In our setup, the YOLOv2 network \cite{yolov2}, known for real 
time object detection and  localisation, is used to learn the semi-lexical tokens. 
The training set is prepared with images from Caltech256~\cite{caltech256}, 
VOC~\cite{voc2012}, and consists of only {\em100} images of bicycles. 

The semi-lexical tokens in a given image containing any of the three objects are 
identified using the same network and tagged as $Tag(\mathcal{T}_t)$ = $(name, pos)$, where 
$name$ refers to the name of the component and $pos$ refers to the bounding box 
containing the component. An example of identified components is shown in  
Figure~\ref{fig2:a}, where we consider only semi-lexical tokens for {\em wheel} and 
{\em frame}. The tagged components decide the truth of the predicates {\em wheel} and 
{\em frame}, for example if the network identifies one wheel $w_1$ and a bicycle frame 
$f$ the predicates $wheel(w_1)$ and $frame(f)$ are set to true. The 
$inrange$ predicate is set to true if the euclidean 
distance between the identified components lie within permitted range, the range check 
also ensures that the identified components are unique. For bicycles, $range = 
[min_{distance},max_{distance}]$ between two components is calculated over the training 
dataset $\mathcal{T}$. Distance between component $c_1$ and $c_2$ of the $i^{th}$ instance 
$distance_i = \sqrt{\{(c_{1x}-c_{2x})/w\}^2 + \{(c_{1y}-c_{2y})/h\}^2}$ where $w$ and $h$ are the 
width and height of $image_i$.

If the network is unable to identify all the components required for logical deduction in
the first pass (for example, if only one wheel of a bicycle was identified), then we mask
the identified components and reduce the threshold by $\epsilon = 0.1$ and continue 
searching for the required parts until the component is found or threshold $\geq 0.2$. 
Drawing parallel from the semi lexical integrity constraints formalised for handwritten 
digits in Section~3 the reduced threshold search enforces {\em reasoning assisted similarity constraint}, trying to look for other components of an object in the pictures if some 
supporting component for the object is found. After the object is identified to belong 
to a particular class, we check for similarity between two similar types of components using Equation \ref{eq1:b} to enforce {\em reasoning assisted dissimilarity constraint}. If the two components are not similar, 
they are tagged to be inconsistent.  For example, in Figure~\ref{fig2:c}, one of the 
wheels belong to a motorcycle. Even though the rules are satisfied, this wheel will not
be tagged as a part of the bicycle.

% The recognition methodology is summarised in Algorithm \ref{algo:overall}.
% \setlength{\textfloatsep}{1pt}
% \begin{algorithm}[t]
% \SetAlgoLined
% \DontPrintSemicolon
% \caption{Pseudo code for identifying cycle}
% \label{algo:overall}
% \KwIn{$image$}
% \SetKwFunction{FMain}{Main}
% \SetKwProg{Fn}{Function}{:}{}
% \Fn{\FMain{$image$}}{
% $components(name, position)$ $\gets run\_YOLOv2(image)$\;
% \uIf{$BicycleLogic(components)$}{
%    \KwRet{$Bicycle$}\;
% }
% \uIf{$UnicycleLogic(components)$}{
%    \KwRet{$Unicycle$}\;
% }
% \uIf{$TicycleLogic(components)$}{
%    \KwRet{$Tricycle$}\;
% }
% \KwRet{$None$}\;
% }
% \end{algorithm}

A semi-lexical analysis reduces the burden on pure machine learning. For example, the 
traditional YOLOv2 network used for detecting complete objects uses $9$ convolutional
layers. In our setup, we need to identify the components rather than objects, and
therefore we use a smaller network with substantially less training data. Based on the 
performance of different plots in Figure~\ref{fig2:b} we chose a network with $7$ 
layers. The proposed bicycle detection methodology is tested with clear bicycle images from 
Caltech256 and WSID-100~\cite{wsid100} data sets. The algorithms are tested with 
unicycle and tricycle images as well, which do not require any extra learning because
the components are the same. The results obtained are shown in Table~\ref{tab:2}.
\begin{table*}[ht]
\centering
\ra{1.0}
\setlength\tabcolsep{3pt}
\resizebox{\textwidth}{!}{
\begin{tabular}{@{}llrrrrrrr@{}}\toprule
& & \multicolumn{4}{c}{Accuracy (\%)} \\
\cmidrule{3-6}
Methodology & Hyperparameters & \begin{tabular}[r]{@{}r@{}}WSID-100\\(500 images)\end{tabular} & \begin{tabular}[r]{@{}r@{}}Caltech256\\(165 images)\end{tabular} & \begin{tabular}[r]{@{}r@{}}Uni-/Tricycle\\(150 images)\end{tabular} & \begin{tabular}[r]{@{}r@{}}Not cycle\\(500 images)\end{tabular} & \begin{tabular}[r]{@{}r@{}}Precision\\(\%)\end{tabular} & \begin{tabular}[r]{@{}r@{}}Recall\\(\%)\end{tabular} & \begin{tabular}[r]{@{}r@{}}F1 score\\(\%)\end{tabular}\\ \midrule

\multirow{2}{*}{Our} & ep = 70, OT= 0.40 & 94.40 & 92.12 & 74 & 100 & 100 & 93.83 & 96.81\\
% & 70 & 0.45 & 85.50 & 81.11 & 72 & 100 & 100 & 84.81 & 91.78\\
& ep = 100, OT= 0.40 & 94.40 & 93.93 & 67.33 & 100 & 100 & 94.28 & 97.05\\
%& 100 & 0.45 & 90.20 & 87.77 & 56.36 & 100 & 100 & 89.83 & 94.64\\

YOLOv2 & ep = 100, OT= 0.30 & 77.40 & 73.93 & - & 100 & 100 & 76.54 & 86.71 \\
AlexNet (FR) & $\eta $ = $10^{-5}$ & 86.60 & 81.81 & - & 91.8 & 93.26 & 85.41 & 89.16 \\
VGG16 (FR) & $\eta $ = $10^{-5}$ & 91.20 & 88.48 & - & 85.6 & 89.31 & 90.52 & 89.91 \\
%AlexNet (CR) & $\alpha $ = $10^{-5}$ & 100 & 100 & - & 100 & 100 & 100 & 100 \\
VGG16 (CR) & $\eta $ = $10^{-5}$ & 100 & 100 & - & 85.6 & 90.23 & 100 & 94.86 \\
VGG16 (KNN) & KD tree, k = 2 & 86 & 76.36 & - & 61 & 74.03 & 83.60 & 78.53 \\
VGG16 (OCS) & $\gamma$ = 0.004, $\nu$ = 0.15 & 72.60 & 62.42 & - & 66.20 & 73.39 & 70.08 & 71.69 \\
\midrule
\multicolumn{9}{l}{\tiny{ep = epochs \quad OT = Objectness Threshold \quad FR = Fully Retrained \quad CR = Classifier Retrained \quad OCS = One Class SVM classifier  \quad KNN = K Nearest Neighbours classifier}} \\
\bottomrule
\end{tabular}}
\caption{Comparison of proposed detection methodology with some standard classification networks. All the networks are trained with only 100 bicycle images except the classifier retrained networks which are trained on Imagenet. Precision, Recall and F1-scores are calculated only on the bicycle data sets.}
\label{tab:2}
\end{table*}

Table~\ref{tab:2} illustrates that semi-lexical deduction outperforms standard CNN based identification techniques in terms of F1 score, that is, our model maintains good precision recall balance in all cases when tested on different bicycle data-sets. Though the VGG16 network with only classifier retrained layer has better accuracy, its feature extraction layers are trained on Imagenet dataset \cite{imagenet} and the network miss classifies objects like tennis racket, cannon, etc. as bicycle lacking in precision. Our method has the added advantage of low data requirement (trained on only 100 bicycles), explainability in terms of choice of tokens that trigger the final classification outcome and detecting classes of objects sharing similar components without training.

\section{Related Work}
\label{sec5}
CNN's have shown exceptional performance in computer vision 
tasks like image recognition, object localization, segmentation, etc.~\cite{7780459,rcnn,yolov2}. Unfortunately, CNN's lack interpretability, which is necessary for learning complex scenarios in a transparent way, and are known to fail in simple logical tasks such as learning a transitive relation~\cite{mathrule}. These networks are also susceptible to adversarial attacks~\cite{adv1,adv2} and are bad at retaining spatial information~\cite{pool}. Such 
weaknesses occur as the network latches onto certain high dimensional components for pattern matching~\cite{adversaries}. Another major drawback that deep learning faces is the  requirement of huge amounts of annotated data. 

Hence, a lot of current research advocates merging the power of both connection based and symbol-based  AI ~\cite{symbolic,reasoning,evans:2018,satnet}. These works aim at solving problems using a SAT optimization formulation. However, the methods are limited by their memory requirements. Other advances, like neuro-symbolic concept learner, proposes hybrid neuro-symbolic systems that use both AI systems and Neural Networks to solve visual question answering problems~\cite{neuro}, have the advantage of exploiting structured language prior.

For computer vision tasks symbolic formulation of image grammar has been explored using stochastic AND-OR  graphs that are probabilistic graphical models that aim to learn the hierarchical knowledge semantics hidden inside an image~\cite{stcarg}. The parse graph generated from a learnt attribute graph grammar is traversed in a top-down/bottom-up manner to generate inferences while maximizing a Bayesian posterior probability. This method requires a large number of training examples to learn the probability distribution. Also, the graph can have exponentially large number of different topologies. Methods that use pure symbolic reasoning for identification, like ellipse and triangle detection for bicycle identification ~\cite{bicycle}, do not generalize well. Works by ~\cite{bpl} learn concepts in terms of simple probabilistic programs which are built compositionally from simpler primitives. These programs use hierarchical priors that are modified with experience and are used as generative models rather than identification. Also, ~\cite{tlt} uses special prototypical layers at the end of the model that learns small parts called prototypes from the training image. The test image is then broken into parts and checked for similarity against the learnt prototype parts and prediction is made based on a weighted combination of the similarity scores. In general, the methods discussed do not account for ambiguous tokens that can exhibit overlapping membership in multiple classes.

\section{Conclusions}
\label{sec6}
The real world often presents itself in wide diversity, and capturing such 
diversity purely in symbolic form is not practical. Therefore, inherent in our
ability to interpret the real world is a mapping from the non-lexical artifacts
that we see and the lexical artifacts that we use in our reasoning.
Semi-lexical languages, as we propose in this paper, provides the formal basis
for such reasoning. For implementing this notion on real world problems in
computer vision, we use machine learning (ML) to learn the association between the
non-lexical real world and the alphabet of the formal language used
in the underlying reasoning system. An important difference with related
work is that the ML-based interpretation of the real world is assisted by the 
reasoning system through the similarity / dissimilarity consistency 
constraints.

\balance
\bibliographystyle{named}
\bibliography{ijcai20}

\end{document}